\documentclass[10pt,twocolumn,letterpaper]{article}

\usepackage{iccv}
\usepackage{times}
\usepackage{epsfig}
\usepackage{graphicx}
\usepackage{amsmath}
\usepackage{amssymb}

\usepackage[breaklinks=true,bookmarks=false]{hyperref}

\iccvfinalcopy 

\def\iccvPaperID{****} 

\ificcvfinal\fi

\begin{document}

\title{Geometry aware 3D generation from in-the-wild images in ImageNet}

\author{Qijia Shen\\
University of Oxford\\
{\tt\small shenqijia11@gmail.com}
\and
Guangrun Wang\\
University of Oxford\\
{\tt\small wanggrun@gmail.com}
}

\maketitle
\ificcvfinal\thispagestyle{empty}\fi

\begin{abstract}
   Generating accurate 3D models is a challenging problem that traditionally requires explicit learning from 3D datasets using supervised learning. Although recent advances have shown promise in learning 3D models from 2D images, these methods often rely on well-structured datasets with multi-view images of each instance or camera pose information. Furthermore, these datasets usually contain clean backgrounds with simple shapes, making them expensive to acquire and hard to generalize, which limits the applicability of these methods. To overcome these limitations, we propose a method for reconstructing 3D geometry from the diverse and unstructured Imagenet dataset without camera pose information. We use an efficient triplane representation to learn 3D models from 2D images and modify the architecture of the generator backbone based on StyleGAN2 to adapt to the highly diverse dataset. To prevent mode collapse and improve the training stability on diverse data, we propose to use multi-view discrimination. The trained generator can produce class-conditional 3D models as well as renderings from arbitrary viewpoints. The class-conditional generation results demonstrate significant improvement over the current state-of-the-art method. Additionally, using PTI, we can efficiently reconstruct the whole 3D geometry from single-view images.
\end{abstract}

\section{Introduction}

\begin{figure}[h]
  \centering
  \includegraphics[width=\linewidth]{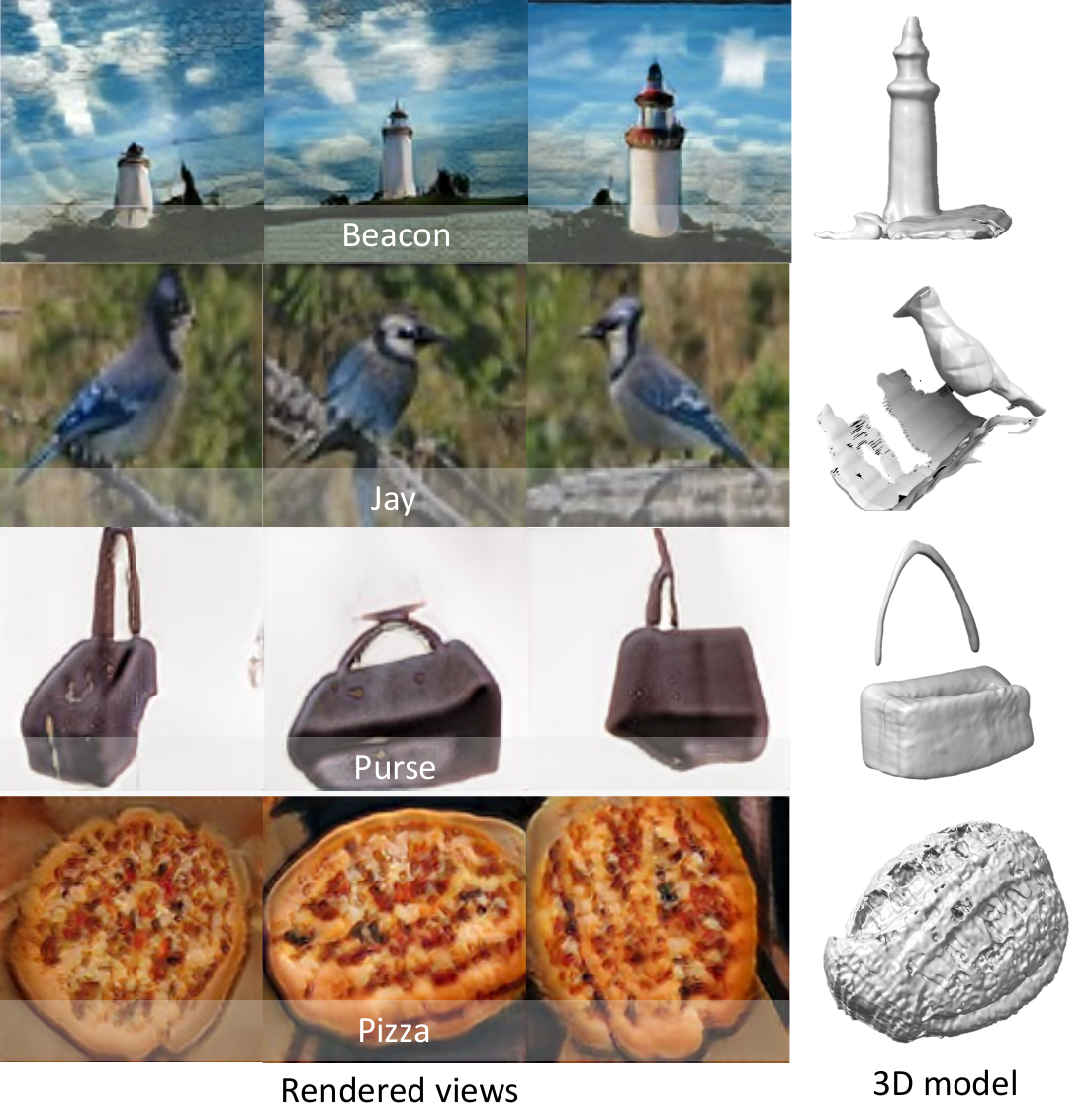}
  \caption{Selected examples of 3D models and synthesized views respectively. Using in-the-wild images on Imagenet for training, our method successfully learned a class-conditional 3D generative model.}
  \label{fig:example}
\end{figure}
Generative adversarial networks (GANs) have been extensively studied for generating high-quality 2D images. Methods such as StyleGAN and its improved versions  \cite{karras2019style,karras2020training,karras2020analyzing} are capable of generating 2D images with fine details. Other works, such as BigGAN  \cite{brock2018large} and StyleGAN-XL  \cite{sauer2022stylegan}, have further extended the capabilities of GANs to large and diverse datasets like ImageNet  \cite{russakovsky2015imagenet,deng2009imagenet}. Although techniques have been developed to enable some limited capability of rotating the viewpoints of the generated images, these methods do not explicitly model the 3D geometry.

Although 3D generation has been a branch of research for a long time, it has received less attention than 2D generation. Early 3D generation methods heavily relied on 3D convolutional neural networks (CNNs) \cite{wu2016learning} and 3D datasets for supervision, but obtaining 3D data is challenging. To address this, \cite{tulsiani2017multi} proposed using multi-view images as supervision for 3D model reconstruction. This concept was further extended by \cite{tulsiani2018multi}, who jointly estimated pose and shape by aligning the directions of similar objects, eliminating the need for explicit 3D data. Recent works on 3D geometry-aware synthesis, such as pi-GAN \cite{chan2021pi}, GRAF \cite{schwarz2020graf}, and GIRAFFE \cite{niemeyer2021giraffe}, have developed the idea of 2D supervision and eliminated the need for 3D CNNs. These methods are based on 2D CNNs but extract 3D information using representations such as NeRF and make the model aware of 3D geometry. As a result, 2D-based methods have reduced the computational requirements for training 3D generative networks.

Despite these advancements, the development of 3D generation still lags behind 2D generation. While large models for 2D generation, such as CLIP and DALL-E \cite{radford2021learning,ramesh2021zero}, have been developed to learn from billions of images collected from various websites, 3D generative methods are still confined to limited datasets consisting of either human faces or a single object class with rendered images against a clean background, such as ShapeNet \cite{shapenet2015} and CompCars \cite{yang2015large}. This is because current 3D generative methods still rely, to some degree, on unique 3D prior information such as camera pose and multiple views. As a result, 3D generative methods cannot take full advantage of the rich 2D image resources available.

However, we believe that learn from in-the-wild images in diverse datasets without any additional information other than anything required for learning on 2D is crucial for 3D generation to benefit from large datasets. As humans, we do not calculate exact pitch and yaw angles relative to the object, nor do we observe objects without background. However, after viewing enough scenes containing objects of the same category, we can naturally imagine a new instance in the same category or complete views different from the given one. Therefore, we believe that the ability to generate diverse and complex 3D models from in-the-wild images without any additional information is a critical step towards unlocking the full potential of 3D generation.

Therefore, in this work, our aim is to learn a 3D generative model from the Imagenet-1k dataset  \cite{deng2009imagenet,russakovsky2015imagenet}, with the model being class-conditional to generate specific categories. However, we face the challenge of not having the camera pose of each image, which we couldn't estimate during the preprocessing stage, as was done in previous works on face datasets  \cite{chan2022efficient}. To address this, we sample camera poses for each rendering during training from a fixed radius sphere uniformly. However, this approach may result in views that are not present in the dataset, leading to instability in training the generator. Therefore, we propose a method of simultaneously rendering multiple views for discrimination and calculating the average loss. The camera poses for these views are placed apart to ensure that at least some views capture directions that exist in the dataset. Additionally, the demand for the generator and discriminator is imbalanced in the baseline method (EG3D), as generation is on 3D while discrimination is on 2D. To overcome this, we propose modifying the architecture of the generator backbone and decoder to increase their model capacity.

Our evaluation mainly focuses on the ImageNet-1k dataset, with an additional evaluation on the ShapeNet Core dataset. Qualitatively, we were able to generate high-quality 3D models and 2D synthesized views for both datasets. Furthermore, an additional benefit of our method is that the directions of all objects within each class are aligned. For quantitative evaluation, we computed metrics such as FID, KID, and IS, and compared our method with the baseline method. Our results show a significant improvement over the baseline method.

In summary, our contributions are the following:
\begin{itemize}
    
    \item We modified the architecture of backbone of StyleGAN2 generator, as well as the decoder for implicit neural representation, to adapt to a larger scale datasets.
    
    \item We got rid of the need for camera pose for training and adopted whole-sphere sampling scheme for camera view points during training.
    
    \item We proposed multi-view discrimination to stablize GAN training and delay mode collapse for generating better 3D models as well as synthesized views. 
    
    \item We successfully learned 3D models from ImageNet without knowledge of camera pose. We demonstrated the results of synthesized multi-view image and 3D models.

\end{itemize}

\section{Related work}
\begin{itemize}
    \item \textbf{Neural representations and rendering.}
    The representation of 3D scenes and objects has always been a challenge due to the intrinsic complexity of 3D data. While 2D images are relatively straightforward to represent, acquiring ground truth for 3D scenes is demanding and explicitly using 3D information during training requires a large amount of memory. Methods have been developed to represent objects and backgrounds separately  \cite{szabo2019unsupervised}, but these are limited to simple scenes.
    
    Recently, implicit neural representations have emerged as a promising method to represent scenes in a continuous and memory-efficient way. Techniques such as using 3D data as supervision for reconstruction of 3D surfaces, optimizing signed distance functions, and using 3D geometry as regularization have been developed to enable this approach.
    
    The recent emergence of the implicit neural representation provides a method to represent scenes in a continuous and memory efficient way. Techniques like using 3D data (e.g., point cloud) as supervision for reconstruction of 3D surface  \cite{atzmon2020sal}, optimizing signed distance functions  \cite{park2019deepsdf, sitzmann2020metasdf}, or using 3D geometry as regularization  \cite{gropp2020implicit} have been developed based on this method. 

    With the development of differentiable neural renderers, explicit 3D supervision is no longer required. Instead, the whole models can be optimized on 2D multi-view or even single-view images \cite{liu2019learning, liu2020dist, niemeyer2020differentiable,sitzmann2019scene,yariv2020multiview, mildenhall2021nerf,saito2019pifu,sitzmann2019scene} using the implicit representation of surface or the whole 3D volumes as an intermediate representation. However, querying the implicit representation is time-consuming as it requires passing through a MLP.
    
    To address this issue, a hybrid representation  \cite{devries2021unconstrained, peng2020convolutional, chan2022efficient} that combines the advantages of explicit and implicit representations has been developed. This approach encodes spatial information explicitly, eliminating the need to query positional information in the implicit representation and significantly reducing the number of layers in the MLP. Recent studies have followed this idea and produced efficient methods for learning 3D generative models.
    
    \item \textbf{Generative 3D-aware image synthesis.}
    Generative Adversarial Nets (GANs) have been studied for a long time, with various techniques developed to target different purposes. BigGAN and BigBiGAN \cite{donahue2019large, brock2018large} focus on conditional generation of diverse data, while StyleGAN and its improved versions \cite{karras2019style, karras2020analyzing, karras2020training, karras2021alias} are geared towards generating photorealistic images and methods for image editing \cite{wang2018high}, domain transfer \cite{zhu2017unpaired}. However, these methods lack awareness of the 3D structures of scenes and objects and can generate images that do not have correct geometry.

    Recently, GANs with 3D awareness have emerged, generating an intermediate representation of 3D structures for rendering different views of 2D images. Prior to the emergence of implicit neural representation, explicit 3D voxel grids \cite{gadelha20173d, zhu2018visual} or meshes \cite{szabo2019unsupervised} were generated first and then rendered into 2D views. However, these methods were limited by the heavy memory overheads of explicit representation and were hard to extend to more complex scenes with higher resolutions.
    
    Recent advances incorporated implicit neural representation into the 2D GANs framework and has achieved the controllable editing  \cite{niemeyer2021giraffe} of multiple objects and more realistic novel view \cite{chan2021pi, chan2022efficient}. Representative works like pi-GAN  \cite{chan2021pi}, GRAF  \cite{schwarz2020graf} and GIRAFFE  \cite{niemeyer2021giraffe} all use neural radiance field directly as the generator. The input is the position and the output is the color and density. Differentiable rendering methods proposed in  \cite{martin2021nerf, barron2021mip} are used to synthesize different views of images for discrimination.

    Lift. StyleGAN  \cite{shi2021lifting} and EG3D  \cite{chan2022efficient} take a slightly different approach. They use StyleGAN as the backbone to generate an intermediate representation which is then decoded and rendered. Additionally, EG3D  \cite{chan2022efficient} proposes an explicit-implicit hybrid representation of 3D volumes that balances memory and rendering efficiency.

    \item \textbf{Learning 3D models from images in the wild}
    Learning 3D structure from images in the wild presents a greater challenge than learning from well-designed datasets like ShapeNet \cite{shapenet2015}. This is because most datasets used for 3D-aware image generation have limited diversity, as they are created from 3D modelling software. Images in the wild are unpaired, unaligned, and can vary in illumination, shape, and color. Additionally, camera pose is unknown, which is an important factor for many current techniques. Some works have attempted to generate models from images in the wild, but these methods heavily rely on handcrafted modelling  \cite{booth20183d} or strong inductive bias such as symmetry  \cite{wu2020unsupervised}. However, recent work NeRF-W by Martin et al.  \cite{martin2021nerf} successfully generated a 3D scene from a collection of in-the-wild pictures, demonstrating the potential of learning from images in the wild.

    \item \textbf{Single-view reconstruction}
    Reconstructing the full 3D shape typically requires multiple views, but it is now possible to perform shape completion based on a single view using some trained networks.  \cite{xiang2022any,zhang2018learning} both first infer the depth image from the input and then complete the whole 3D shape. CodeNeRF \cite{jang2021codenerf} directly learns an implicit neural representation of shape and texture, but their results have only been shown to work on data with limited diversity such as cars and chairs \etc.
    
    A recent work RealFusion \cite{melas2023realfusion} could complete the shapes of arbitrary single image input. It uses pre-trained diffusion models \cite{rombach2022high} as the prior and optimize the text-prompt for generating other views consistent with the input image. However, the reconstruction time for each 3D model could take several hours using multiple GPUs. 

    One application of our work is efficient and simple 3D shape completion. Using pivotal tuning inversion on the trained generator \cite{roich2022pivotal}, we can easily reconstruct the full 3D structure in minutes in an end-to-end manner, without relying on complex intermediate representations. 

\end{itemize}

\section{Methods}
\begin{figure*}
\begin{center}
\includegraphics[width=.9\linewidth]{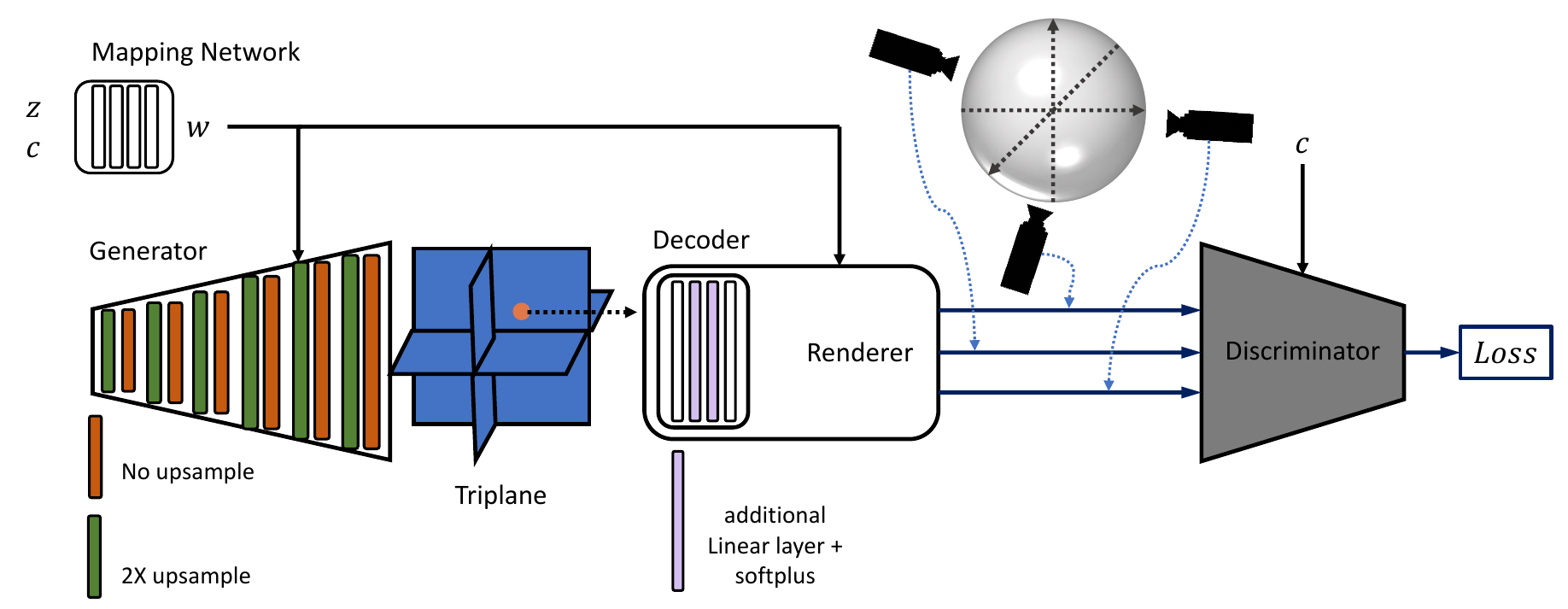}
\end{center}
   \caption{Structure of our method. We add some addtional layers to the StyleGAN2 generator. The original StyleGAN2 is composed of sequential synethesis blocks. Each block doubles the resolution of the feature maps in the previous block, as indicated by the green blocks in the figure. After each original synthesis block, we add an additional synthesis block with same structure, indicated by orange block in the figure, except $2\times$ upsampling. We also increase the depth of the decoder by adding additional linear layers (purple blocks in the figure). Renderer will synthesize three views using the camera poses sampled uniformly from a sphere.}
\label{fig:network}
\end{figure*}
\subsection{Triplane representation}
Generating 3D models from a dataset composed of natural images is a challenging task that typically requires longer training times for convergence. Previous implicit neural representations \cite{martin2021nerf} have been slow in querying and are not suitable for this task. On the other hand, a complete explicit representation would require too much memory and impede training at higher resolutions. Explicit-implicit hybrid representation supports fast querying and small memory overheads. As shown in Fig. \ref{fig:network}, we modified the number of output channels at the last layer to $32\times 3$, with 32 channels for each orthogonal plane. When queried with a point in space, the decoder generates density and color information by taking the summation of interpolated values at that point from all planes as input. Although interpolation could be used, and the final image resolution is no longer determined by the output resolution of the 2D generator, we found that the triplane resolution would largely affect the sophistication of the generated 3D models. Using a low triplane resolution with high image resolution increases the burden of the renderer to beat against the discriminator and does not help improve the geometry of the generated 3D model. Therefore, we continue to use triplanes with a resolution of $256\times 256$ and select EG3D as our baseline method \cite{chan2022efficient}\vspace{-11pt}

\subsection{Scaling generator to a more diverse dataset}
\subsubsection{Modification to decoder}
The original decoder for triplane, which is a two-layer MLP, was designed to be learned specifically for each scene, similar to that of NeRF. However, this decoder is not sufficient for more diverse datasets like ImageNet. To address this issue, we propose to increase the depth of the decoder by adding two intermediate fully connected layers with the same number of input and output features. These additional layers enable the decoder to handle more complex information from triplane when learning in a diverse dataset.\vspace{-11pt}

\subsubsection{Modification to backbone}

The backbone of the baseline model follows the design of StyleGAN2 \cite{karras2020analyzing} which was originally designed to allow detailed control of different characteristics and may be too restrictive for training on a diverse dataset like ImageNet \cite{grigoryev2022and}. On the other hand, success of BigGAN \cite{brock2018large} in modeling ImageNet is largely attributed to its increased model size. Additional comparisons between BigGAN-deep and BigGAN also support this idea.\vspace{-3pt}

However, we found that increasing the model size of only the generator worked better for our 3D generative task. While the original generator and discriminator pairs perfectly for 2D images, they become imbalanced when synthesizing 3D  consistent images, as the task for the backbone now is generating 3D models which is much harder than before. Demands for the discriminator remain similar.

\vspace{-3pt}We conducted experiments by increasing the depth of the discriminator with and without a larger generator, and we observed that mode collapse occurred faster in both cases. This validated our assumption that the discriminator has an easier task and thus outperforms the generator.

To improve the backbone's capacity, we increase the depth of the StyleGAN2 generator. To be more specific, the original StyleGAN2 generator has $N=log_2(resolution)$ synthesis blocks. Each synthesis block would upscale the resolution of the feature maps in previous block for $2\times$. We add an addtional synthesis block without upscaling after each original block at different resolutions, indicated by orange blocks shown in fig. \ref{fig:network}. This modification allows increasing the network depth but not changing the overall design. The total number of training parameters of the generator is doubled after modification.

\subsection{Introducing multi-view discrimination}

As mentioned earlier, there is an imbalance in the demands placed on the generator and discriminator which can easily lead to mode collapse. Even when using an increased size generator model, signs of mode collapse are still inevitable after a period of training. One obvious indicator of mode collapse is the lack of diversity in the colors of the synthesized images.

What poses even greater challenge for 3D generation is that the distribution of camera pose in the dataset is unknown. Although camera poses during generation are sampled uniformly from a fixed radius sphere, in-the-wild datasets are unlikely to have uniform coverage of all angles. If images are rendered from a viewpoint that has not been covered in the dataset, the discriminator would inaccurately adjust the generator, leading to increased training instability.

To address these challenges, we propose a straightforward idea of using multi-view discrimination. After generating the triplane, we render several views simultaneously and send them all to the discriminator for separate loss calculation, with the final loss being the average of these views. By sampling from the sphere uniformly for multiple views, there will always be at least one or two views that fall into viewpoints that exist in the dataset. This helps stabilize training of the generator.

To enable longer training of the generator, improve the fidelity of 3D models generated, and minimize loss of training efficiency, we introduce multi-view discrimination just before the start of mode collapse to stabilize training and delay the onset of mode collapse.



As shown in fig. \ref{fig:multiple_views}, our assumption for training stabilization stands when the camera poses for multiple views are separated apart. We also allow some perturbation to the uniformly sampled camera poses to prevent the "billboards" effect, where objects appear flat and lack depth when viewed from certain angles. This approach helps to capture a more diverse set of viewpoints and improve the robustness of the model.

\subsection{3D reconstruction from single-view}
To enable 3D reconstruction from a single-view input, we employed the pivotal tuning inversion \cite{roich2022pivotal} technique. The overall inversion is separated into two stages, as illustrated in Figure \ref{fig:pti}. In the first stage, we freeze the weights of the generator and renderer and focus on optimizing the latent vector $w$. To initialize $w$, we use a pre-trained classification network to classify the category to which the real/target image most likely belongs, and predict the class condition $c$. We then input $c$ and a random vector $z$ into the mapping network to obtain the output $w$, which is used as the starting value for the latent optimization process. We calculate the loss function as the Euclidean distance between the generated image and the target image in feature space, using a VGG16 \cite{simonyan2014very} feature extraction network. In the second stage, we freeze the latent vector $w$ while optimizing the parameters of the generator and renderer. The loss function is calculated as the sum of Euclidean distance in both image space and feature space.\vspace{-11pt}

\begin{figure}
  \centering
  \includegraphics[width=\linewidth]{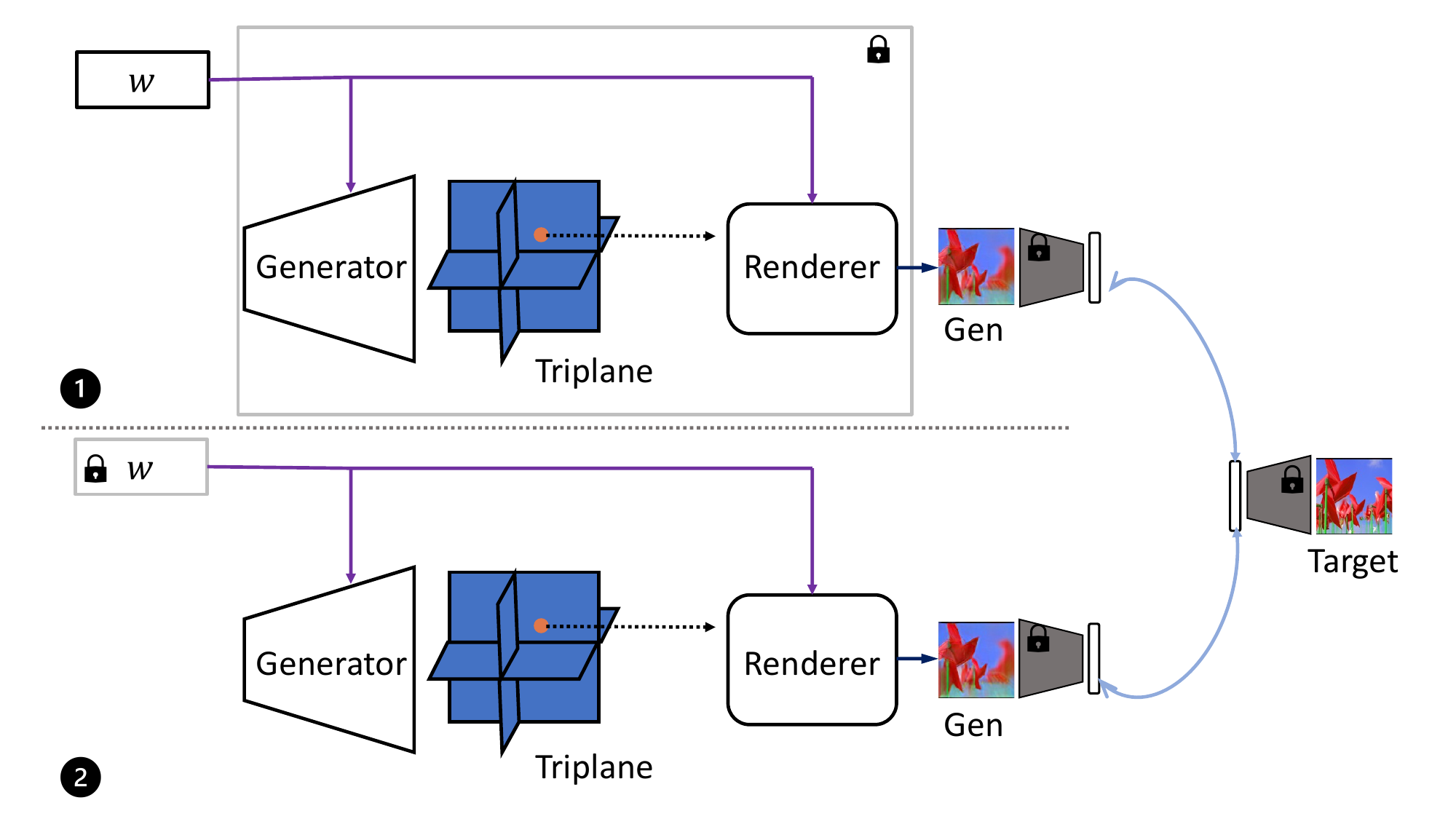}
  \caption{Single-view image inversion is separated into two stages. Latent vector is optimized in the first stage and the generator is optimized in the second stage. The optimization goal is to minimize the distance between the generated image and the target image in both feature space and image space. The camera pose can be arbitrarily chosen, as long as it remains the same during the whole inversion process.\vspace{-11pt}}
  \label{fig:inversion}
\end{figure}

\section{Experiments and results}
\subsection{Datasets}
Our 3D generative model was trained on a subset of the whole ImageNet-1k  \cite{russakovsky2015imagenet} dataset. While ImageNet-1k comprises multiple categories with similar characteristics, our focus is not on detailed discrimination between the similar categories. Instead, we aim to demonstrate a method which is able to learn 3D generation from a diverse set of in-the-wild images. To this end, we selected 100 classes (126,994 images) from ImageNet-1k, spanning a range of subjects from birds and boats to burgers and buildings, with minimal overlap. Most images have complex background, and many contain multiple instances, presenting significant challenges to both the generator and discriminator. 

We also evaluate our model's performance on the ShapeNet  \cite{shapenet2015} which includes 55 classes (52,472 images) and serve as a further validation of our method's generalizability. Training on the ShapeNet also follows the same class-conditional approach and settings as on ImageNet. Each training image is rendered from a different 3D model using Blender and the camera poses for rendering are randomly sampled from a sphere with a fixed radius of 1.2. Notably, the camera poses were not provided to the discriminator during training.

\subsection{Results on ImageNet}
\begin{figure*}
\begin{center}
\includegraphics[width=.9\linewidth]{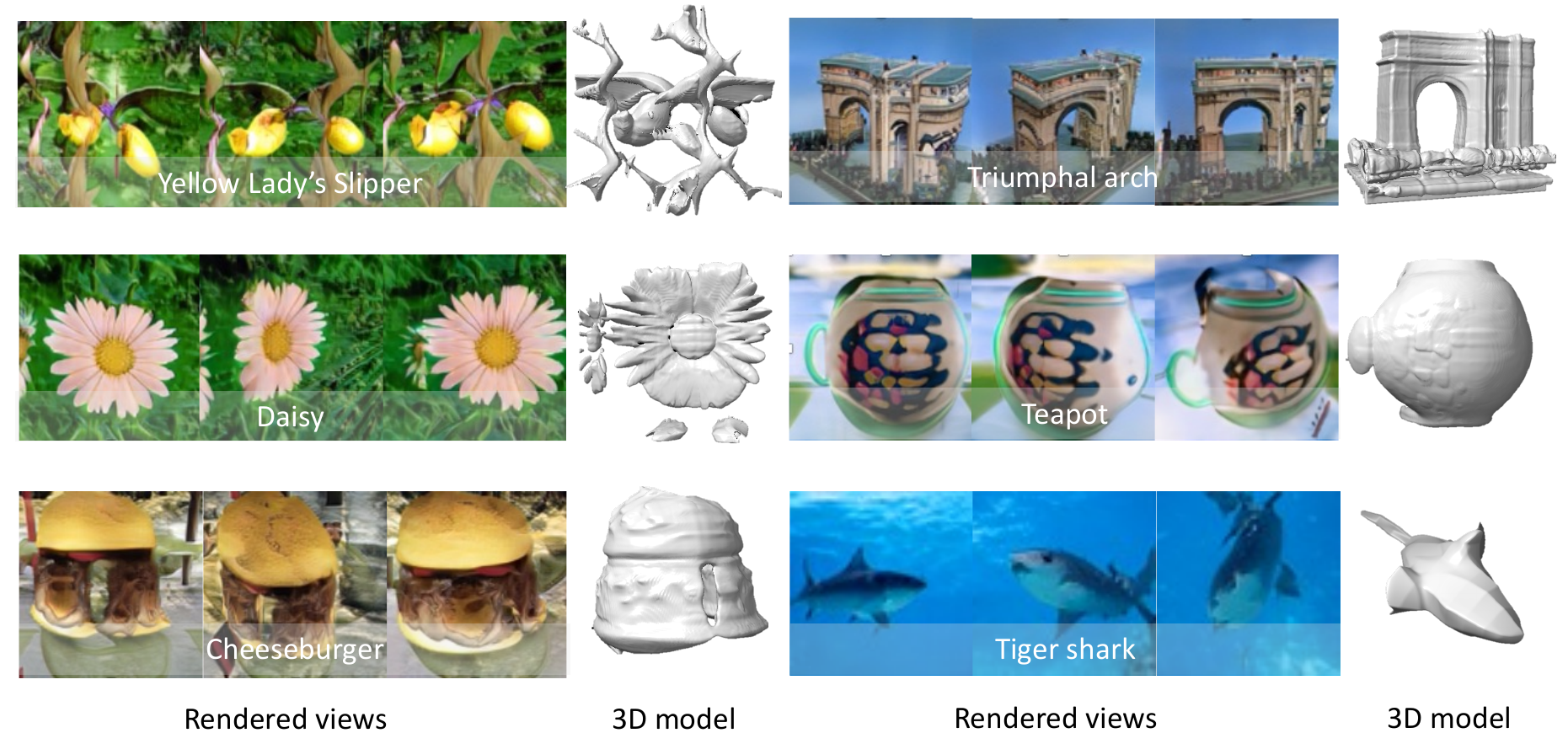}
\end{center}
   \caption{We present selected examples of 3D models generated using our method, along with three different views synthesized from each corresponding model. In this demonstration, we showcase six examples spanning from plants to architecture and everyday objects. These categories represent challenging cases for 3D generative models in the training set, with the Yellow Lady's Slipper having a highly complex background, the daisy having a limited view with a fixed angle, and the cheeseburger being deformed. However, our method has successfully learned the 3D models of these objects, demonstrating its potential and effectiveness.\vspace{-11pt}}
\label{fig:multiple_views}
\end{figure*}

Fig. \ref{fig:multiple_views} showcases examples of 3D models generated with different class conditions, demonstrating the ability of our method to generate 3D models directly, allowing examination from arbitrary viewpoints. Notably, our method learns to align the directions of objects in the same class, occupying most of the field of view, indicating that it learned 3D positions and directions while learning the geometry. Additionally, the model learns to focus on relevant objects given the class condition and adds reasonable backgrounds according to different classes.

Training on a large dataset also has the benefit of inferring reasonable shapes for objects with limited views. For example, in fig. \ref{fig:example}, the 3D geometry of pizza cannot be learned when trained only on pizza data, as all images in the class of pizza have a consistent top view, providing no information about its geometry. However, training with other data allows our model to infer reasonable novel views never seen before.

\begin{figure*}
\begin{center}
\includegraphics[width=.8\linewidth]{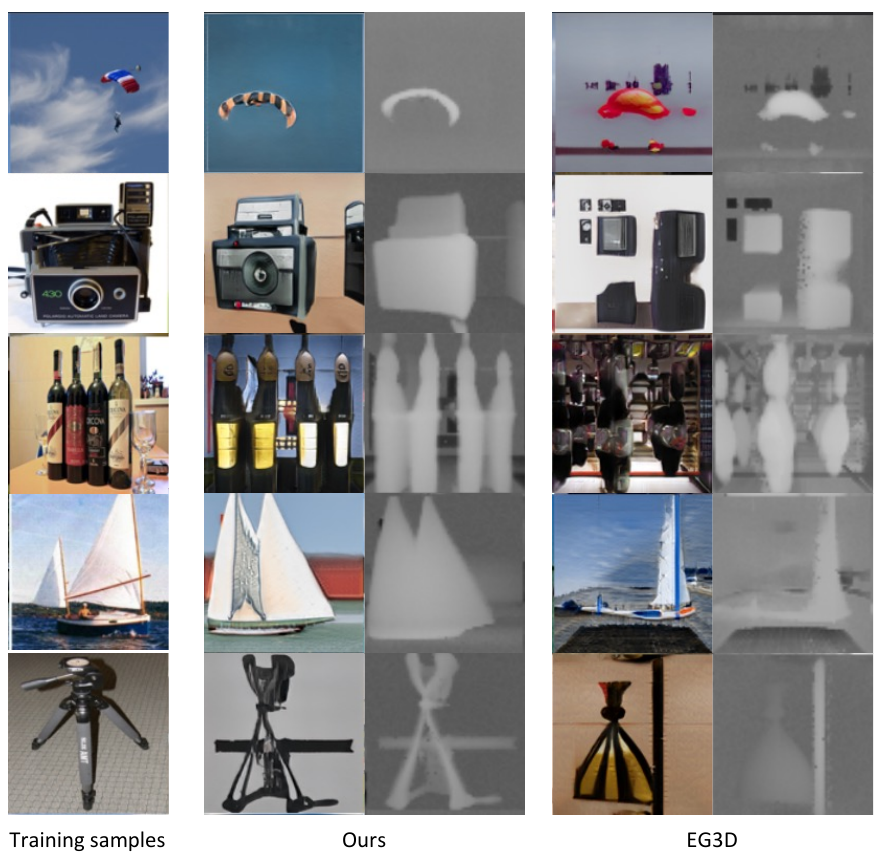}
\end{center}
   \caption{Comparision of samples generated by EG3D and ours. Five classes from top to bottom are parachute, Polaroid camera, wine bottle, yawl and tripod. Our method demonstrates a superior ability to learn the 3D shape of various categories, and it excels at handling challenging cases where multiple instances are presented in the same image. This is particularly evident in the images on the third row, which showcase our method's capability to handle such cases.\vspace{-11pt}}
\label{fig:comparison}
\end{figure*}

Fig. \ref{fig:comparison} provides a visual comparison between the generated samples of our method and EG3D. While EG3D is able to synthesize images that follow the class condition, its quality is inferior and it fails to capture the 3D geometry of the objects. For instance, in the fifth row of fig. \ref{fig:comparison}, which belongs to the category of tripods, EG3D generated images with a four-foot stand, but the depth image shows that it actually generates a "billboard" to deceive the discriminator. It is a common failure when learning 3D-aware generative model and it is also described in the original EG3D paper \cite{chan2022efficient}. They solved the issue by applying random perturbation to the camera pose condition for discriminator, whereas, in our case, the discriminator is not aware of the camera poses, and the camera poses for rendered views are randomly sampled. Therefore, the failure of EG3D in this case is mostly attributed to the lack of model capacity.

We evaluated the performance of our generative model on the ImageNet dataset using three metrics, including Fréchet Inception Distance (FID) \cite{eslami2018neural}, Kernel Inception Distance (KID) \cite{binkowski2018demystifying}, and Inception Score (IS) \cite{salimans2016improved}. As shown in Table \ref{tab:generation_comparison}, our method achieved significant improvements over the baseline method EG3D in all three metrics, which is consistent with qualitative comparison between two methods in fig. \ref{fig:comparison}.

It should be noted that the values of these metrics are higher than what we typically see in 2D generation, due to the fact that the camera poses during evaluation are also uniformly sampled from a sphere, which may lead to some views that do not exist in the dataset. Therefore, while it is fair to compare the relative performance between different methods using the same camera pose sampling strategy, the average deviation from the dataset will surely increase. 

We did not include other methods in our comparison because they either perform inferior to EG3D or do not generate 3D models, but only geometry-aware images that do not meet our requirements for direct 3D generation.\vspace{-11pt}

\begin{table}
\begin{center}
\begin{tabular}{|l|c|c|c|}
\hline
Method &FID $\downarrow$ & IS$\uparrow$ & KID $\downarrow$ \\
\hline\hline
EG3D &70.93&11.57 &0.051\\
Ours &47.06&14.23 &0.038\\
\hline
\end{tabular}
\end{center}
\caption{Comparison of generation diversity and fidelity on ImageNet. Our method is significantly superior to EG3D in every aspect.\vspace{-11pt}}
\label{tab:generation_comparison}
\end{table}

\subsection{Ablation study}

\begin{table}
\begin{center}
\begin{tabular}{|l|c|c|}
\hline
Method & FID $\downarrow$ & KID $\downarrow$ \\
\hline\hline
Baseline &70.93 & 0.051\\
+ d & 63.42 &0.047 \\
+ g + d & 53.10&0.041 \\
+ g + d + multi-view & 47.06 &0.038 \\
\hline
\end{tabular}
\end{center}
\caption{Ablation study. "d": modified decoder. "g": modified generator backbone. "multi-view": multi-view discrimination. Modifying the generator and decoder allows the model to learn from diverse dataset. Multi-view discrimination stabilizes GAN training and avoids mode collapse.\vspace{-11pt}}
\label{tab:ablation}
\end{table}

We chose EG3D as our baseline, as it combines the efficiency of triplane with the ability of StyleGAN2 to generate high-quality images. The only difference is that we removed camera pose conditioning for the discriminator.

Table \ref{tab:ablation} indicates that the modified generator backbone is essential for effective learning from the ImageNet dataset. While the modified decoder is also effective, its improvement is restricted when used in isolation. The original generator backbone generates triplanes with limited variability, despite its ability to decode additional features. Combining the modified generator backbone and the decoder can maximize the potential for learning on diverse datasets, as the large generator backbone can encode greater variability in triplanes, while the modified decoder can decode richer information in the triplane.

Comparing the fourth row with the third row in Table \ref{tab:ablation}, we notice the additional boost from using multi-view discrimination. This technique stabilizes the training process and reduces the mode collapse significantly. Therefore, generator could be trained longer and generate 3D models with higher quality. Multi-view discrimination is also effective when resuming training without multi-view discrimination. As the technique does not change the network structure but takes slightly longer for each iteration. Maximum training efficiency could be achieved by starting off training without multi-view discrimination and switching to multi-view discrimination when mode collapse occur.

\subsection{Single-view 3D reconstruction}
\begin{figure}
  \centering
  \includegraphics[width=\linewidth]{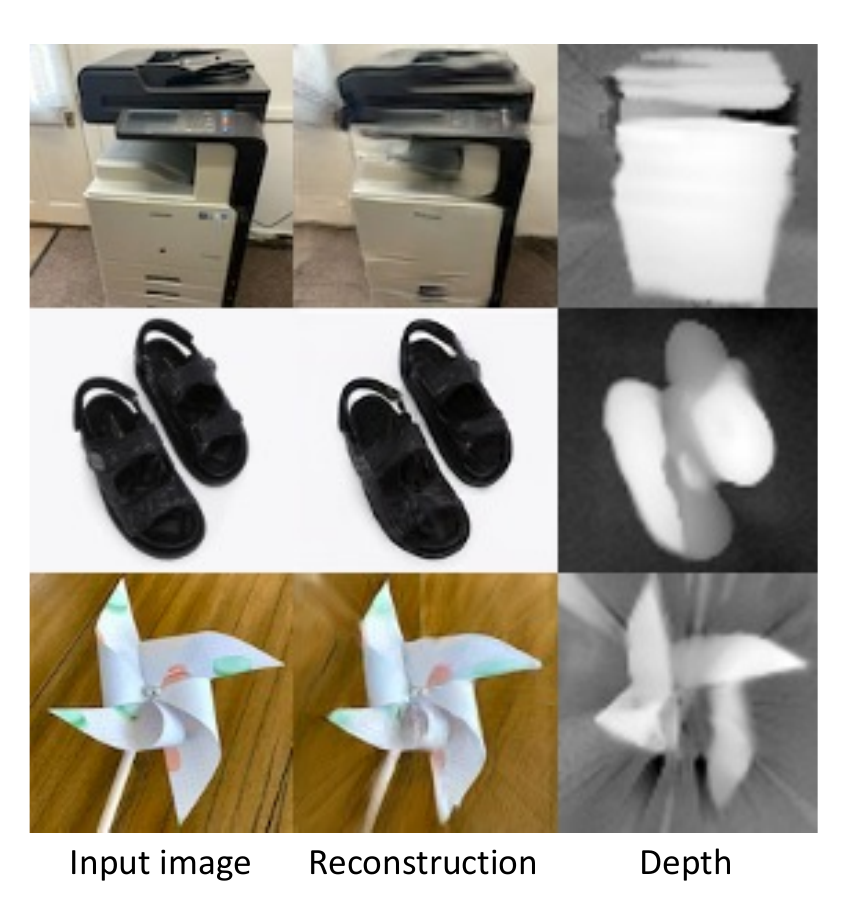}
  \caption{3D shape completion for single-view images collected from Internet.}
  \label{fig:pti}
\end{figure}

One important application for trained 3D generative model is enabling efficient single-view 3D reconstruction. To evaluate our method, we used the label of each category on ImageNet to search for online images as our input. With the help of Pivotal Tuning Inversion \cite{roich2022pivotal}, we were able to complete the 3D shape consistent with the input image, as shown in fig. \ref{fig:pti}. 

We also quantitatively evaluate the reconstruction quality using depth accuracy \cite{chan2022efficient} and LPIPS \cite{zhang2018unreasonable}, where LPIPS measures the perceptual consistency between the input and reconstructed image, and depth accuracy reflects the quality of the generated 3D shape. The pseudo ground truth map was predicted using Dense Prediction Transformer  \cite{ranftl2021vision}. Table \ref{tab:PTI} presents the results for reconstruction using EG3D and ours. 

Our method and EG3D achieved comparable LPIPS scores, as PTI involves the step of fine-tuning the generator, which could easily overfit the input image. However, our method outperforms EG3D in terms of depth accuracy, which heavily relies on previous training on large datasets to complete the correct shape, a point where our method is superior to EG3D.

\begin{table}
\begin{center}
\begin{tabular}{|l|c|c|}
\hline
Method & LPIPS$\downarrow$ & Depth accuracy$\downarrow$ \\
\hline\hline
EG3D &0.23 &0.28\\
Ours & 0.22 & 0.21\\
\hline
\end{tabular}
\end{center}
\caption{Comparison of the quality of single-view image inversion. \vspace{-11pt}}
\label{tab:PTI}
\end{table}

\subsection{Results on ShapeNet}
The ShapeNet dataset \cite{shapenet2015} has also been widely used for training 3D-related tasks. We used the ShapeNet Core V2, which consists of 55 different classes. Previous research in the field generally trains their models on a single class, such as cars, and has not attempted to generate class-conditional samples using the whole dataset. In this work, we also evaluated our method on this dataset.

Table \ref{tab:shapenet} shows that EG3D works well on ShapeNet, as the diversity of ShapeNet is much less than ImageNet. However, our method still outperforms EG3D on ShapeNet. Furthermore, EG3D inevitably shows signs of mode collapse at the final period of training, when generated samples of different classes gradually have similar color and shape. In contrast, our method adopts multi-view discrimination and avoids mode collapse, enabling longer training and producing samples with higher diversity and fidelity.

\begin{table}
\begin{center}
\begin{tabular}{|l|c|c|c|}
\hline
Method & FID $\downarrow$ & IS$\uparrow$ & KID$\downarrow$ \\
\hline\hline
EG3D &19.14 &7.002 &0.0064\\
Ours &12.65 &7.34 &0.0039\\
\hline
\end{tabular}
\end{center}
\caption{Evaluation of the generation fidelity and diversity on ShapeNet. Our method is also significantly better than EG3D.\vspace{-11pt}}
\label{tab:shapenet}
\end{table}

\subsection{Discussion and conclusion}

\textbf{Limitation} Our work still has some limitations as we have not yet fully solved the issue for unknown pose for in-the-wild images. However, based on the characteristic that objects in the same class are aligned in direction, we may explore pose prediction during training. We also have not yet taken advantage of the depth information during training, which could later be used to enforce consistency between RGB and depth information as additional supervision for geometry. We also have not yet explored more architectures other than StyleGAN, but a less restrictive architecture may be more suitable for large scale and diverse data.

\textbf{Conclusion} By modifying the architectures of the generator and applying multi-view discrimination training strategy, we successfully learned a class-conditional 3D generative model from in-the-wild images on ImageNet. Our work presents a promising step towards large scale 3D generation.

{\small
\bibliographystyle{ieee_fullname}
\bibliography{egbib}
}

\end{document}